\documentclass[conference]{IEEEtran}
\usepackage[latin1]{inputenc}
\ifCLASSINFOpdf
  \usepackage[pdftex]{graphicx}
\else
  \usepackage[dvips]{graphicx}
\fi
\ifCLASSOPTIONcompsoc
  \usepackage[caption=false,font=normalsize,labelfont=sf,textfont=sf]{subfig}
\else
  \usepackage[caption=false,font=footnotesize]{subfig}
\fi
\usepackage{url}
\usepackage{soul}
\begin{document}

\title{A Non-Technical Survey on Deep Convolutional Neural Network Architectures}

\author{
\IEEEauthorblockN{Felix Altenberger}
\IEEEauthorblockA{Technical University of Munich\\
85748 Garching, Germany\\
Email: felix.altenberger@tum.de}
\and
\IEEEauthorblockN{Claus Lenz}
\IEEEauthorblockA{Cognition Factory GmbH\\
80797 Munich, Germany\\
Email: lenz@cognitionfactory.de}
}

\maketitle

\begin{abstract}
Artificial neural networks have recently shown great results in many disciplines and a variety of applications, including natural language understanding, speech processing, games and image data generation. One particular application in which the strong performance of artificial neural networks was demonstrated is the recognition of objects in images, where deep convolutional neural networks are commonly applied. 
In this survey, we give a comprehensive introduction to this topic (object recognition with deep convolutional neural networks), with a strong focus on the evolution of network architectures. Therefore, we aim to compress the most important concepts in this field in a simple and non-technical manner to allow for future researchers to have a quick general understanding.

This work is structured as follows: 
\begin{enumerate}
\item We will explain the basic ideas of (convolutional) neural networks and deep learning and examine their usage for three object recognition tasks: image classification, object localization and object detection.
\item We give a review on the evolution of deep convolutional neural networks by providing an extensive overview of the most important network architectures presented in chronological order of their appearances. 
\end{enumerate}
\end{abstract}


\section{Introduction}\label{sec:introduction}
During the last years, artificial agents have been increasingly able to outperform humans in a variety of challenges across many different domains \cite{tesauro1995temporal,schaeffer2007checkers,Atari,bowling2015heads,AlphaGo}. While it seems that computers have obvious advantages over humans in many areas, such as calculus or industrial assembly, they have very recently also managed to outperform humans on tasks that are comparatively easy for humans while being utterly complex problems for artificial agents.

Recognizing faces in an image is a good example for such a task, as every human is able to do it within the fraction of a second. In computer vision, face recognition is a very difficult challenge, where a lot of research is still being conducted. Until recently, artificial agents were not able to achieve results comparable to those of humans, even with the most advanced approaches and the best available hardware. The methodology that finally allowed computer vision to outperform humans on object recognition tasks \cite{ILSVRC} is the \emph{Deep Convolutional Neural Network (DCNN)}, which we will inspect more closely in this survey. Almost 20 years ago, \cite{LeNet5} already proposed the LeNet, a novel DCNN architecture for object recognition, but only in 2012 an implementation by \cite{AlexNet}, the AlexNet, was first able to beat more traditional geometrical approaches on the most popular object recognition contest - the ILSVRC \cite{ILSVRC}. Ever since, DCNNs have been achieving state-of-the-art results on object recognition tasks.

This paper will give a detailed overview of the evolution of DCNN architectures and how they are applied to object recognition challenges. The paper is structured as follows: In Sect. \ref{sec:dcnn}, we will take a closer look at the deep convolutional neural network and how it works. Afterwards, in Sect. \ref{sec:applications}, we will inspect how DCNNs are used for three different object recognition tasks: classification, localization and detection. In Sect. \ref{sec:architectures}, the most influential DCNN architectures, including the LeNet and AlexNet we mentioned earlier, are presented in chronological order and explained. Finally, in Sect. \ref{sec:conclusion}, we will sum up the key aspects covered in this paper and list selected resources for further research.

\section{Deep Convolutional Neural Networks}\label{sec:dcnn}
In this section the basics of deep convolutional neural networks will be explained. Therefore, general artificial neural networks and deep learning will be introduced first, before diving deeper into convolutional layers and operations.

\subsection{Artificial Neural Networks}\label{subsec:ann}
Nature has always been a source of inspiration for scientific advances \cite{Biomimetics}. When given the task of designing an algorithm for object recognition without having much prior knowledge of the field, one might attempt replicating some object recognition system that can be found in nature. Within the human brain, the neocortex is responsible for recognizing very high-level patterns, such as abstract concepts or complicated implications, which is performed by around 20 billion small processing units, called \emph{Neurons}, that are connected with each other and organized hierarchically \cite{Neuroscience,Kurzweil}.

In the field of artificial intelligence, \emph{Artificial Neural Networks (ANNs)} have been built in a way that mimics their biological counterparts, but the reproduction of this processing functionality of the brain has been made with abstractions and major simplifications. Some of the biological details have been omitted either for simplification and computational cost reduction or because of the lack of knowledge about their role. 

\begin{figure}
\includegraphics{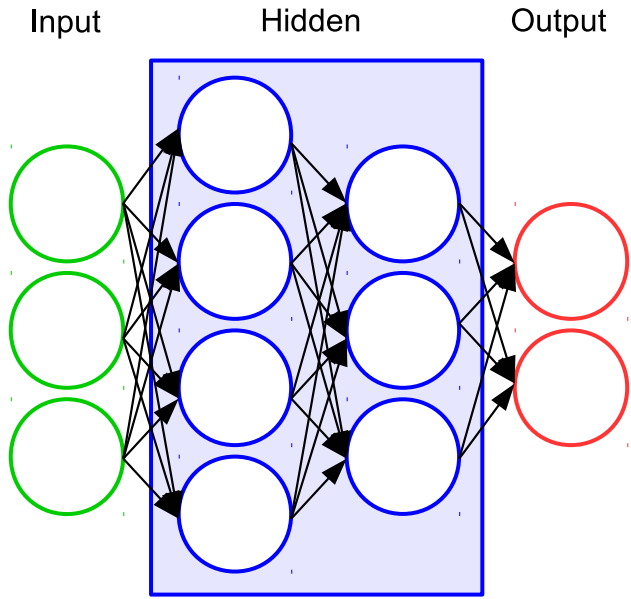}
\caption{A simple artificial neural network, consisting of an input layer, an output layer and two hidden layers}
\label{fig:ann}
\end{figure}

On a high-level perspective, artificial neural networks can be divided into \emph{Layers}. We can distinguish three types of layers: the \emph{Input Layer} and the \emph{Output Layer}, which are representations of the input and output respectively, as well as optional \emph{Hidden Layers}. Hidden layers are simply all the other layers in between, which are performing complementary computations, resulting in intermediate feature representation of the input. A simple example of such an architecture with two hidden layers can be seen in Fig. \ref{fig:ann}.

Neurons that belong to the same layer of a network do generally recognize similar patterns. In the first hidden layers, which are close to the input layer, \emph{Low Level} features like lines and edges can be recognized. As for deeper layers, they are supposed to find \emph{High Level} patterns, e.g. eyes, noses and mouths for the task of face recognition. In the following, we will also use the terms \emph{low level} and \emph{high level} to describe the location of neurons within a network.

In general, ANNs can be divided into two main categories. In the first one, which is called \emph{Feedforward Neural Networks (FFNN)}, neurons are only forwarding their output to neurons of subsequent layers, as shown in Fig. \ref{fig:ann}. The second type of ANN is called \emph{Recurrent Neural Network (RNN)}, where neurons can also transmit information to other neurons within the same layer, to neurons of previous layers and even to themselves. For object recognition purposes, FFNNs are usually used. Thus, RNNs will not be further elaborated on in this paper.

Furthermore, the neurons of a given layer in Fig. \ref{fig:ann} are always connected to all neurons of the previous and subsequent layers. This is the most basic kind of layer, called \emph{Fully-Connected Layer}, which can be used for many different tasks across many different domains. As we will see later, there also exist other types of layers, which are better suited for certain challenges.

In an artificial neural network, connections from one neuron to another are called \emph{Synapses}, which each contain a \emph{Weight}. This weight determines how important the result of the lower level neuron is for the outcome of the higher level neuron. Recognizing eyes might, for instance, be an important prerequisite for recognizing a face, so the corresponding weight should be high. In addition to that, each neuron contains a \emph{Bias} that reflects how likely it is in general that the corresponding pattern is present. The values of these weights and biases are the \emph{Parameters} of the network, which will be learned during the training process. Other attributes of the training procedure exist and are manually chosen. They are referred to as \emph{Hyperparameters}. 

The output of a neuron is a real value. This value is obtained by first performing a linear combination of the inputs, which are the outputs of the previous layer, with the corresponding weights and bias, as shown in the following equation:

\begin{equation}\label{eq:output}
output = (\sum_i{input_i * weight_i}) + bias
\end{equation}

Afterwards, an \emph{Activation Function} is applied to the result, which is enabling the artificial neural network to also approximate very complex functions by performing non-linear transformations. An example of such an activation function is the \emph{Rectified Linear Unit (ReLU)} \cite{ReLU}, which simply discards negative values, as shown by Equation \ref{eq:relu}, and is very popular due to its simplicity and effectiveness in practice \cite{AlexNet}.
\begin{equation}\label{eq:relu}
f(x) = max(x,0)
\end{equation}

\subsection{Deep Learning}\label{subsec:dl}
In order to train a network, a long repetitive learning procedure is applied. At each iteration of the training, a \emph{Loss Function}\footnote{A comparison of popular loss functions has been made by \cite{Loss}} is evaluated. It determines the network's prediction quality and is dependent on the type of training. For instance, in \emph{Unsupervised Learning}, the loss aims to put an emphasis on constraints that the network should have e.g. ability to reconstruct its input, as performed in \emph{Autoencoders} \cite{Autoencoder}. In \emph{Supervised Learning}, where pairs of inputs and corresponding target outputs are available, the loss can be defined as the measure of similarity between the network's predictions and the target outputs. For object recognition, supervised learning is usually used, for which reason it will be the only learning type considered in the remainder of this paper.

In both cases, a \emph{Parameter Update Scheme} is used to alter the network parameters with respect to the defined loss. Most of these update schemes are based on computing the gradient of the loss function with respect to the model parameters and transmitting the updates from the output layer to earlier layers with a method named \emph{Backpropagation}.

When the term \emph{Deep} is used to describe a network, it refers to the number of layers that comprise the network. The term \emph{Deep Learning} is describing the procedure of performing machine learning tasks with deep artificial neural networks \cite{DLbook}. In reality, the best performing deep neural networks are nowadays consisting of hundreds of layers. Since it is often hard to understand what a specific neuron is recognizing, it is difficult to tell how or why a given deep neural network is working (or not). Another challenge of using deep architectures is that different layers might learn and adapt at a different pace. Especially the earlier hidden layers, which are close to the input layer, do frequently either learn much slower (caused by very low gradients) or much faster (caused by high gradients) also leading to an oscillatory behavior. These two problems are referred to as \emph{Vanishing Gradient Problem} and \emph{Exploding Gradient Problem} respectively \cite{NNDLbook}. These can nowadays be prevented quite well by a variety of techniques, including gradient norm clipping \cite{pascanu2012understanding},  proper network parameter initialization, proper choice of activation functions, as well as input normalization \cite{hochreiter1998vanishing,pascanu2013difficulty}.

\subsection{Convolutional Layers and CNNs}\label{subsec:cnn}

When tackling deep learning tasks, it is generally recommended to train and run the models on raw inputs, without manually extracting any features before. The reason for this is that a network trained on the raw input could learn to extract these features on its own, but in contrast to working with prebuilt features, it would also be able to further optimize the feature extraction as the network improves. If the input is an image, it would, therefore, be desirable to work directly with its raw pixel values. Since an image consists of many pixels and each pixel is possibly represented by multiple color values, the representation of that image in the input layer can become highly complex. A full HD RGB image with $1920\times1080$ pixels would, for instance, require an input layer consisting of about six million neurons. If one would use the simple fully-connected network architecture described in Sect. \ref{subsec:ann}, each neuron in the subsequent layer would then be connected to about six million neurons and if the first fully-connected layer would contain just $1000$ neurons, the total number of parameters would amount to over six billion. Since the network has to optimize all of these parameters, the training process could then become very time and storage intensive. 

In order to solve this computational problem, a different kind of network architecture is used, called \emph{Convolutional Neural Network (CNN)}. CNNs are specifically designed for working with images. For this reason, the neurons of a layer are organized across the three dimensions, height, width and depth, just like the pixels in an image where the depth dimension would differentiate the different color values. In addition to that, CNNs introduce two new types of hidden layers in which each neuron is only connected to a small subset of the neurons in the previous layer, to prevent the aforementioned problem. How these layers work and what they are used for will be explained in the following.

\subsubsection{Convolutional Layer}\label{subsubsec:convolution}
In order to deal with the problems that fully-connected layers faced when processing images, \emph{Convolutional Layers} are used in CNNs instead. What separates a convolutional layer from a fully-connected one is that each neuron is only connected to a small, local subset of the neurons in the previous layer, which is a square sized region across the height and width dimensions. The size of this square is a hyperparameter named \emph{Receptive Field}. For the depth dimension, there is no hyperparameter that has to be defined, as the convolutions are by default always performed across the whole depth. The reason for this is that the depth dimension of the input does typically define the different colors of the image and it is usually necessary to combine them in order to extract any useful information.

Neurons of the convolution operator can recognize certain local patterns of the previous layer's output. Since the patterns that are recognized should be independent of their position in the image, all neurons will be forced to recognize the same pattern by making all of them share one single set of parameters. This concept is referred to as \emph{Parameter Sharing}. In order to now recognize multiple different features within one layer, it is required to have several \emph{Filters}, where each filter is a group of neurons that recognize a certain pattern at different locations in the image. In the convolutional layer, the depth dimension is then specifying to which filter a given neuron belongs.

\begin{figure}
\includegraphics{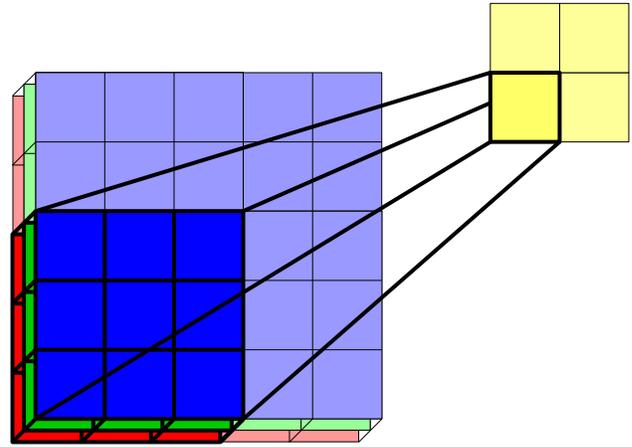}
\caption{A neuron of a convolutional layer performing a convolution operation with a $3 \times 3$ receptive field on an RGB image}
\label{fig:conv}
\end{figure}

Another reason for why the convolution operations are performed across all depth values is that neurons in a convolutional layer, which are stacked on top of others, should have their features jointly considered in the next layer. A neuron in a convolutional layer will, therefore, be connected to $r * d$ neurons of the underlying layer, where $r$ is the size of the receptive field and $d$ is the depth of the previous layer. When performing a convolution directly on the input layer of an RGB image with a receptive field of $3\times3$, each neuron in the layer would, for instance, be connected to $27$ input neurons, consisting of a $3\times3$ square of pixels with three neurons per pixel, as illustrated in Fig. \ref{fig:conv}.

\begin{figure}
\includegraphics{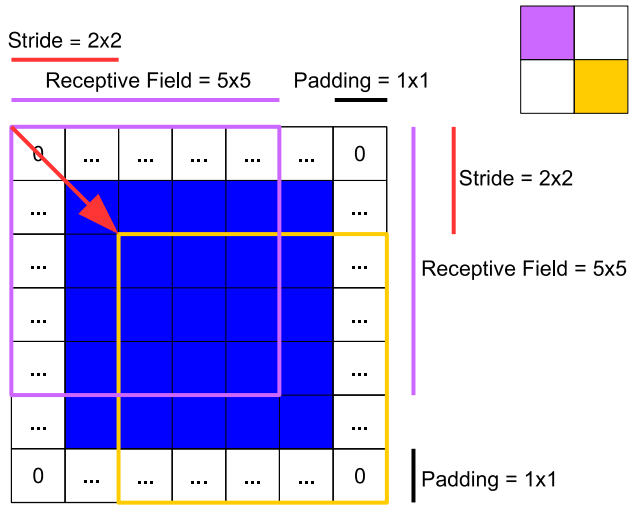}
\caption{A visualization of the different hyperparameters of convolutional layers: receptive field, stride and padding}
\label{fig:conv_hyperparameters}
\end{figure}

How many convolutions are being conducted is defined by another hyperparameter, called \emph{Stride}, which determines how big the gap between two scanned regions is.
Without using any further hyperparameters we would always perform fewer convolutions on inputs close to the borders. Adjusting the \emph{Padding} hyperparameter can make that more even, as it adds an additional border of $0$ values around the original input.
Another reason for why padding is applied in some implementations is to make the convolution result have a certain width and height, e.g. making the output have the same size as the input. A visualization of the three hyperparameters can be seen in Fig. \ref{fig:conv_hyperparameters} and the output size of a convolutional layer can be calculated by the following equation:

\begin{equation}\label{eq:conv_out}
out = \frac{in - receptive field + 2 * padding}{stride} +1
\end{equation}

\subsubsection{Pooling Layer}
The third kind of layer, which has the purpose of decreasing the complexity of CNNs, is the \emph{Pooling Layer}. Similarly to the convolutional layer, neurons in the pooling layer are connected to a square sized region across the width and height dimensions of the previous layer. The main difference between convolution and pooling is that a pooling layer is not \emph{Parametrized}\footnote{The only parametrized layers we will inspect in the scope of this paper are the fully-connected and the convolutional layers.}. This means that neurons in the pooling layer do not have weights or biases that will be learned during the training process but instead perform some fixed function on its inputs. Additionally, the pooling operation does not combine neurons with different depth values. Instead, the resulting pooling layer will have the same depth as the previous layer and it will only combine local regions within a filter.

\begin{figure}
\includegraphics{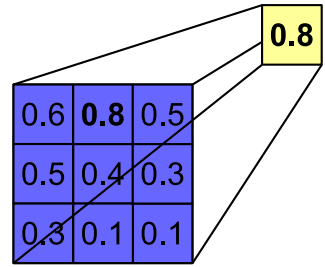}
\caption{A neuron of a pooling layer performing a max pooling operation with a $3 \times 3$ receptive field on neurons of an underlying layer}
\label{fig:pooling}
\end{figure}

One common type of pooling is \emph{Max Pooling}, where the result of combining a number of neurons is the maximum value that any of them returned, which is illustrated in Fig. \ref{fig:pooling}. Since all neurons in the convolutional layer recognize the same pattern, the result of the max pooling operation can be interpreted as whether that pattern has been recognized in the pooling area or not, but the exact location will not be relevant anymore \cite{NNDLbook}. Other pooling variants, such as \emph{Average Pooling}, perform a different function but can be interpreted similarly.

How many neurons are combined across each of the two dimensions and how large the gap between two pooling operations should be is again defined by the receptive field and stride hyperparameters that were used for defining convolutional layers.

\section{Applications of DCNNs for Object Recognition Tasks}\label{sec:applications}
After having described deep convolutional neural networks in the previous section, we will now turn our focus to how these are used for object recognition purposes. More specifically, we will learn about three object recognition tasks - classification, localization and detection - and how each of them can be tackled with DCNNs.

\subsection{Classification}\label{subsec:classification}
\begin{figure*}
\includegraphics{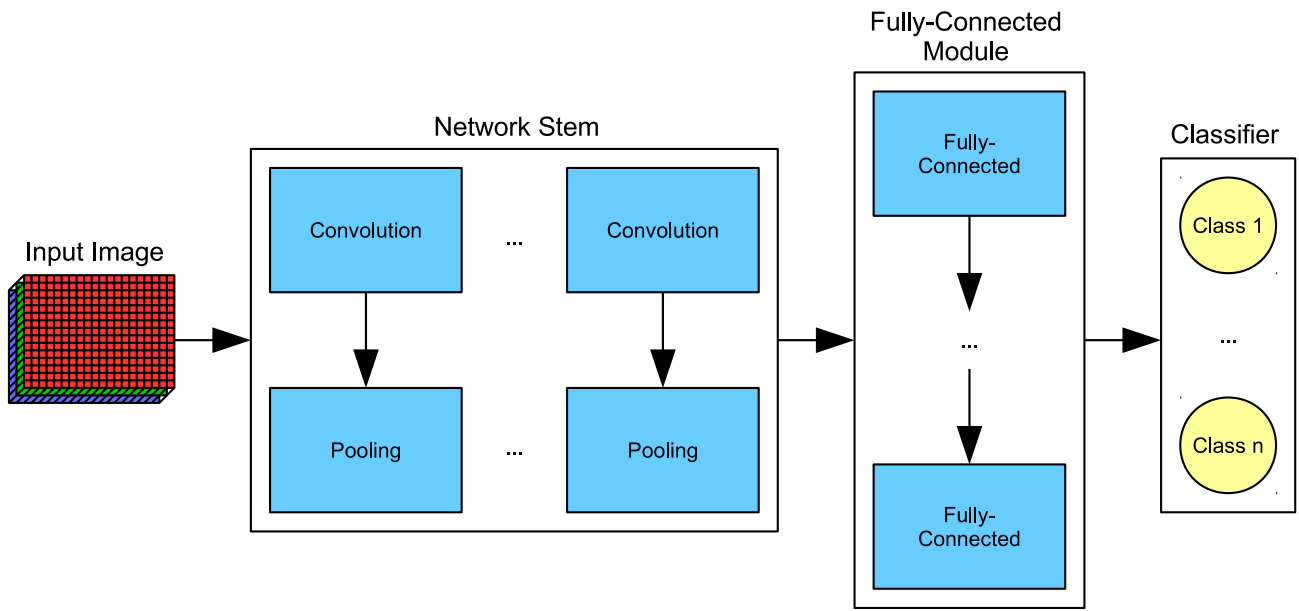}
\caption{The default DCNN architecture for image classification. Multiple convolutional and pooling layers first extract abstract features, which are forwarded to a classifier by fully-connected layers}
\label{fig:classification}
\end{figure*}

The task of \emph{Image Classification} describes the challenge of categorizing a given image into one of several classes. A possible application of this could be the recognition of hand-written digits, where the input image is classified as one of the ten classes. For this task \cite{LeNet5} developed a DCNN architecture in 1998, the \emph{LeNet-5}, which we will inspect more closely later. This network defined the default classification approach, where the DCNN first performs several convolutions and pooling operations in order to extract high-level features. We will refer to this part as \emph{Network Stem} in the remainder. The network stem is then followed by some fully-connected layers, which we call the \emph{Fully-Connected Module}, that is connecting the stem to the output layer. The whole architecture is illustrated in Fig. \ref{fig:classification}.

The output layer for classification tasks consists of one neuron per class and the values of these neurons are representing the score of each class. If we choose a score distribution, where every score is between zero and one and where all class scores add up to one, the values of each neuron can then be interpreted as the probability of whether the class is present.

\subsection{Localization}\label{subsec:localization}
For \emph{Localization}, the information about which category an image belongs to is already available and the task is to instead figure out where exactly the object is located in the image. This location is typically specified by a two-dimensional bounding box, which consists of four values that describe the location of two opposite couples of corners. Finding these four values is the main challenge of localization and is commonly referred to as \emph{Bounding Box Regression}.

To perform a localization task, we can use a similar architecture as the one we defined for classification. The only thing that has to be modified is the final output layer, which can simply be replaced by another output layer that performs the bounding box regression instead.

\begin{figure*}
\includegraphics{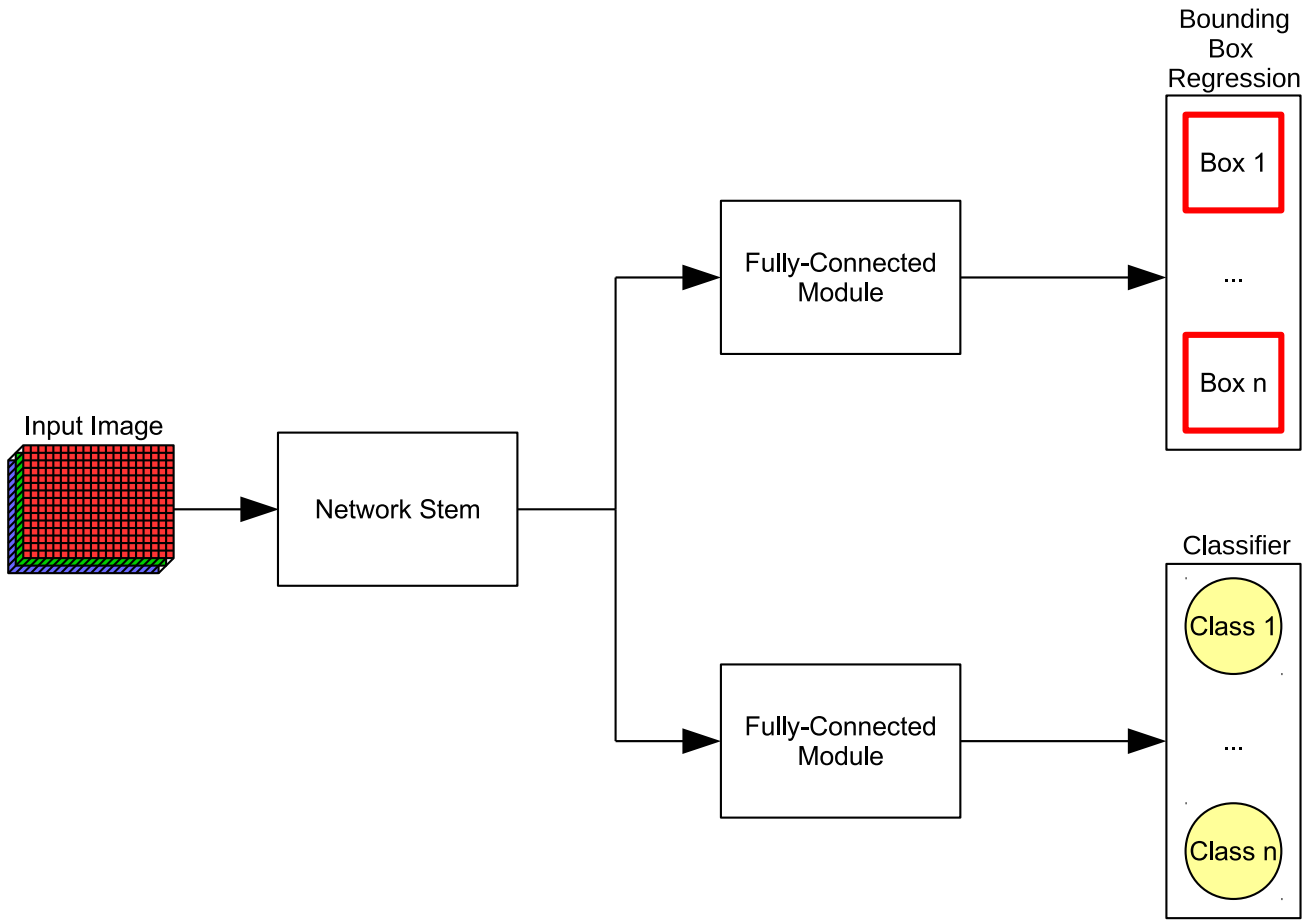}
\caption{A DCNN architecture for multi-class localization. The extracted features of the convolutional and pooling layers in the network stem are used to simultaneously locate the object by bounding box regression and classify it}
\label{fig:localization}
\end{figure*}

Classification and localization can also be combined so that a fixed amount of objects in an image will be classified and also located. This task, called \emph{Multi-Class Localization}, can be tackled with DCNNs by fusing the two architectures that we have seen previously. The resulting architecture will then consist of the network stem, as well as one output layer and corresponding fully-connected layers for each of the two subtasks, which can be seen in Fig. \ref{fig:localization}.

\subsection{Detection}
In contrast to multi-class localization, the number of objects in a given image is not known prior to the execution, when performing \emph{Object Detection}. In order to use DCNNs for such object detection tasks, the architecture needs to be extended to handle the flexible amount of detections. For this purpose, multiple detection methods have been developed, which will be inspected in the following.

\subsubsection{R-CNNs}
One possible way of predicting object detections in an image is by feeding a large number of image parts from the original image into a DCNN, which is then performing multi-class localization to locate and classify the main object in it. Instead of selecting these image parts randomly, \emph{Region-based Convolutional Neural Networks (R-CNNs)} \cite{RCNN} are using \emph{Region Proposal Networks} to only extract potentially interesting regions. These regions are called \emph{Regions of Interest (ROIs)}. They are obtained by running a quick segmentation to spot blob-like structures. 

\cite{FastRCNN} proposed methods for drastically reducing the execution times of R-CNNs, which was achieved by performing only one forward pass through the network for each image and by merging the modules for classification and bounding box regression into one single network.

\begin{figure*}
\includegraphics{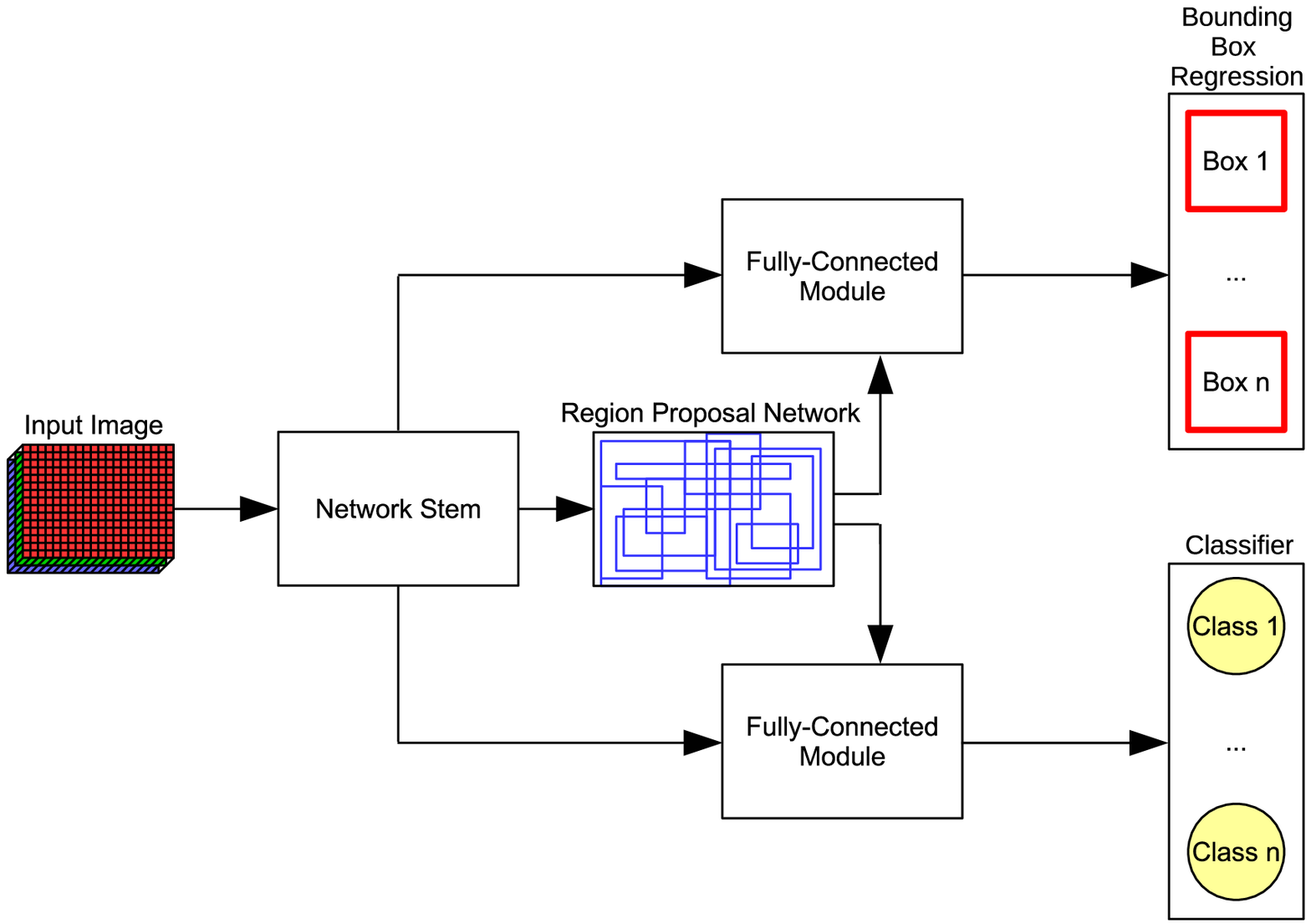}
\caption{A faster R-CNN architecture for object detection. A region proposal network is using the features computed in the network stem to produce ROIs, on which classification and bounding box regression are performed}
\label{fig:fasterrcnn}
\end{figure*}

 \cite{FasterRCNN} managed to further improve the R-CNN execution times by integrating the region proposal network into the remaining network as well, making the resulting \emph{Faster R-CNN} network able to learn end-to-end from the raw pixels of the input image. The structure of such a faster R-CNN model is displayed in Fig. \ref{fig:fasterrcnn}.

\subsubsection{R-FCN}
\begin{figure}
\includegraphics{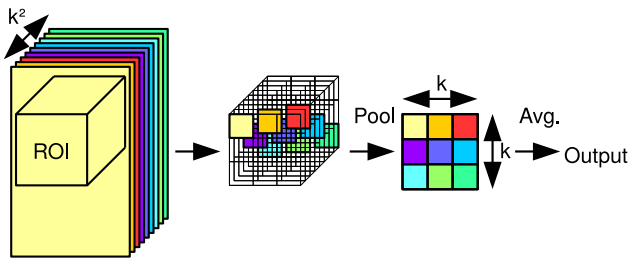}
\caption{A position-sensitive convolutional module of an R-FCN with $k=3$. The first convolutional layer consists of $k^2=9$ depth slices corresponding to certain locations. A given ROI is then split into $k^2=9$ parts and for each part information from the corresponding depth slice is retrieved and pooled. The resulting values are averaged in the end to obtain the final prediction}
\label{fig:rfcn}
\end{figure}

Faster R-CNN architectures are able to achieve strong detection results but are also very complex. Since the fully-connected layers at the end of the network were shown to have a particularly strong impact on the training and execution time, a new ROI-based detection architecture was proposed by \cite{RFCN}, called \emph{Region-based Fully Convolutional Networks (R-FCN)}. These networks are structured very similarly to the faster R-CNN architecture, but instead of using fully-connected modules to predict classes and bounding boxes for each ROI, R-FCNs use \emph{Position-Sensitive Convolutional Modules}. Such a module consists of a convolutional layer with $k^2$ depth slices, called \emph{Score Maps}, where each score map represents the corresponding part of the ROI and computes prediction scores for each class, as well as a pooling layer that is combining the information of the convolutional layer into a $k \times k \times 1$ set of neurons that each correspond to one of the $k^2$ ROI parts and contain prediction scores for all classes. At the end, the scores of the neurons are averaged in order to retrieve the final prediction of the network. The structure of such a position-sensitive convolutional module can be seen in Fig. \ref{fig:rfcn}.

\subsubsection{YOLO}
\begin{figure*}
\includegraphics{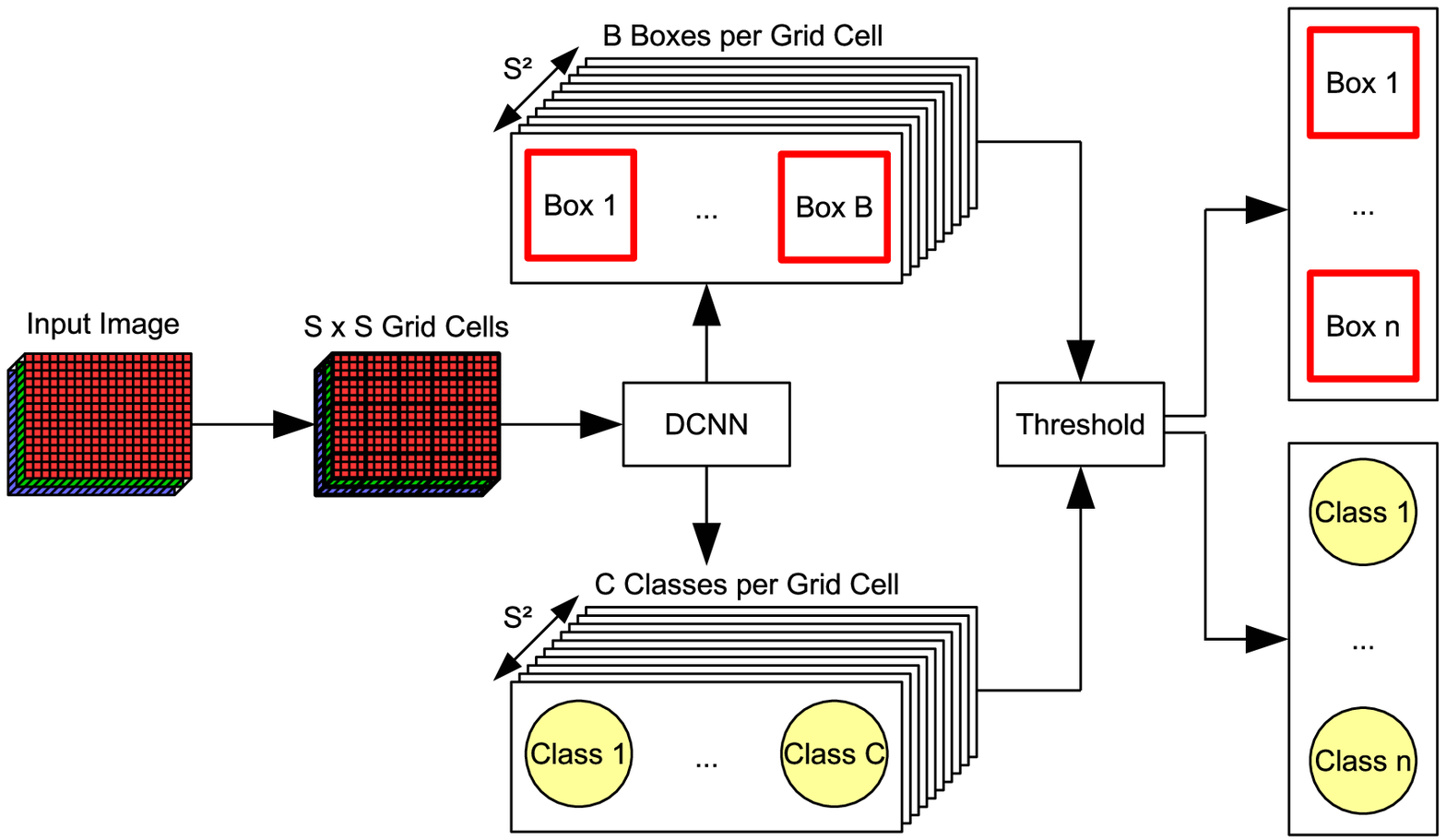}
\caption{The YOLO detection architecture. The input image is overlayed by a grid consisting of $S \times S$ cells. Afterwards, a DCNN is predicting $B$ bounding boxes and $C$ classes per grid cell. The results are thresholded to obtain the final predictions}
\label{fig:yolo}
\end{figure*}

In contrast with region proposal based techniques, \emph{Single-Shot} detection architectures do not predict any regions of interests, but instead, a fixed amount of detections on the image directly, which are then filtered to contain only the actual detections. These networks do therefore have much faster execution times than region-based architectures but are found to also have a lower detection accuracy \cite{RFCN}.
\emph{YOLO}, short for \emph{You Only Look Once}, is a very simple single-shot detection architecture that replaces ROIs by performing a multi-box bounding box regression on the input image directly \cite{YOLO}. In order to do so, the image is overlayed by a grid, and for each grid cell, a fixed amount of detections are predicted, as can be seen in Fig. \ref{fig:yolo}.

Therefore, the whole network has to be evaluated only once per image leading to very fast execution times that are well suited for real-time applications. Additionally, since the model works on whole images it can also make more use out of contextual information. However, since YOLO computes a fixed amount of predictions per region, it is not well suited for tasks where many objects can be located very close to each other and struggles with the detection of objects that have a strong variance in their aspect ratios.

\subsubsection{SSD}
In order to handle the problems of YOLO that arise due to the fixed amount of predictions and cell sizes, \emph{Single Shot MultiBox Detectors (SSD)} have been developed, which predict detections of different scales and also make predictions for multiple different aspect ratios. Therefore, SSD detectors can make finer predictions, leading to significantly better results.

\begin{figure*}
\includegraphics{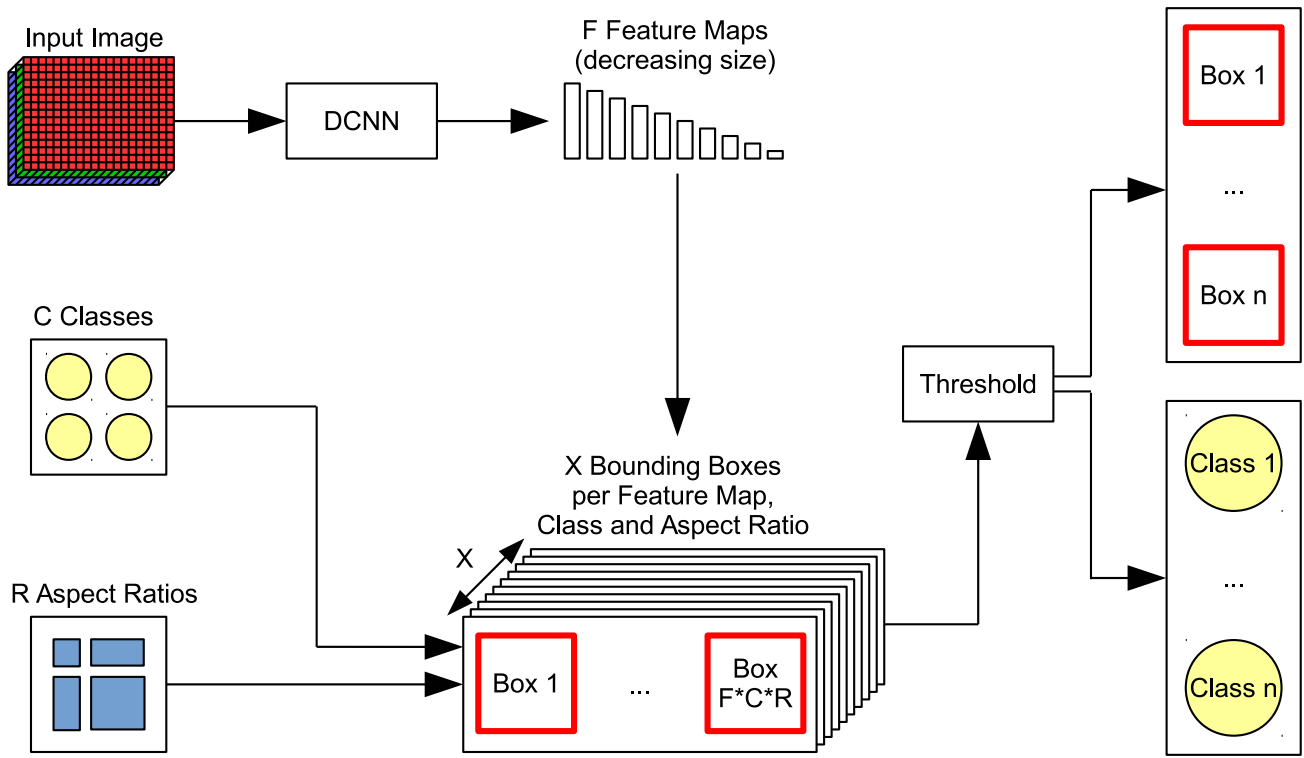}
\caption{The SSD detection architecture. Multiple convolutional layers (feature maps) of decreasing size are appended to the DCNN. For each feature map, a certain amount of detections per class are made with varying aspect ratios and the results are thresholded}
\label{fig:ssd}
\end{figure*}

Similarly to YOLO, the input image is first fed into a convolutional neural network, but instead of performing bounding box regression on the final layer, SSDs append additional convolutional layers that gradually decrease in size. For each of these additional layers, a fixed amount of predictions with diverse aspect ratios are computed, resulting in a large number of predictions that differ heavily across size and aspect ratio. Therefore SSDs are less vulnerable to varying occurrences of objects, leading to significantly better detections than YOLO while preserving a similarly fast execution time \cite{SSD}. An example SSD architecture, differentiating the same amount of aspect ratios as the original paper \cite{SSD} can be seen in Fig. \ref{fig:ssd}.

\subsubsection{YOLOv2}
The basic YOLO was improved on by \cite{YOLO9000}, who released the second version of YOLO, called \emph{YOLOv2}. YOLOv2 contains various improvements in comparison to the first version, such as the ability to predict objects at different resolutions and computing first bounding box predictions by clustering. Additionally, the input size is repeatedly changed to a random value during training, which is enabling YOLOv2 to perform good predictions across various resolutions. As a result, YOLOv2 is able to achieve significantly better detection results than YOLO and was reported \cite{YOLO9000} to have an even better performance than the SSD detector.

\section{DCNN Architectures}\label{sec:architectures}
In this section, we will present the most influential DCNN architectures that have shaped the current state-of-the-art in object recognition. Most of the architectures became famous by winning the \emph{ImageNet Large Scale Visual Recognition Competition (ILSVRC)}\footnote{More information on the ILSVRC can be found here: http://image-net.org/challenges/LSVRC} at some point \cite{ILSVRC}. By inspecting the very best architectures of each year in a chronological order, we will also understand how fast the field is advancing and which trends and new approaches have been developed in each year.

\subsection{LeNet-5 (1998)}\label{subsec:lenet}
\begin{figure*}
\includegraphics{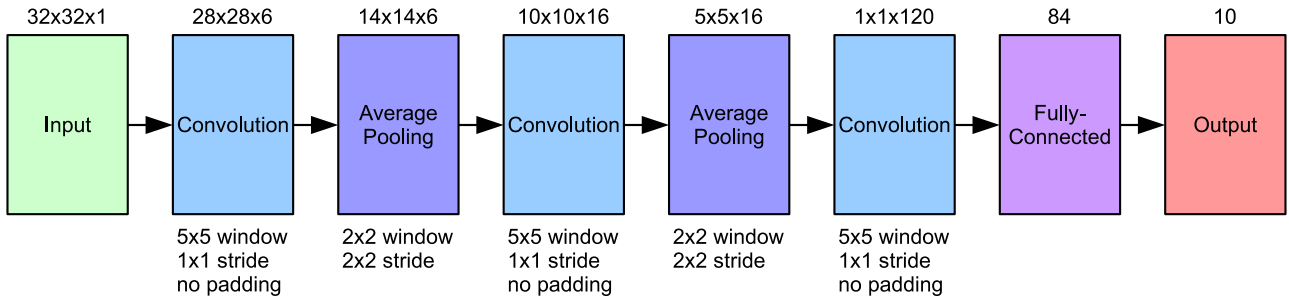}
\caption{The architecture of the LeNet-5. The network consists of an input layer, a convolutional layer, a pooling layer, a second convolutional layer, another pooling layer, another convolutional layer, a fully-connected layer and an output layer from left to right}
\label{fig:lenet}
\end{figure*}

Most of the DCNNs that are being used for object recognition today are based on the basic architecture that was developed by \cite{LeNet5}. This architecture is known as \emph{LeNet-5}, which was used to read digits from $32\times32$ pixel images. The basic architecture of the LeNet-5 can be seen in Fig. \ref{fig:lenet}.

As can be seen, the network architecture is relatively simple, as it only consists of an input layer of size $32\times32$, an output layer of size $10$, as well as three $5\times5$ convolutional, two $2\times2$ pooling and one fully-connected layer in between, making it a total of six hidden layers. Since all convolutional and pooling layers use a stride of one and no padding, the size of each dimension is reduced by $4$ during each convolution and is halved by each pooling operation. The general idea behind the design is to perform multiple convolutions with max pooling between two operations and connecting the final convolutional layer via fully-connected layers to the output layer. This is exactly the default classification architecture that was presented in the previous section and, as we'll show later, this idea has been the basis for most of the other networks in this section as well.

\subsection{AlexNet (2012)}\label{subsec:alexnet}
\begin{figure*}
\includegraphics{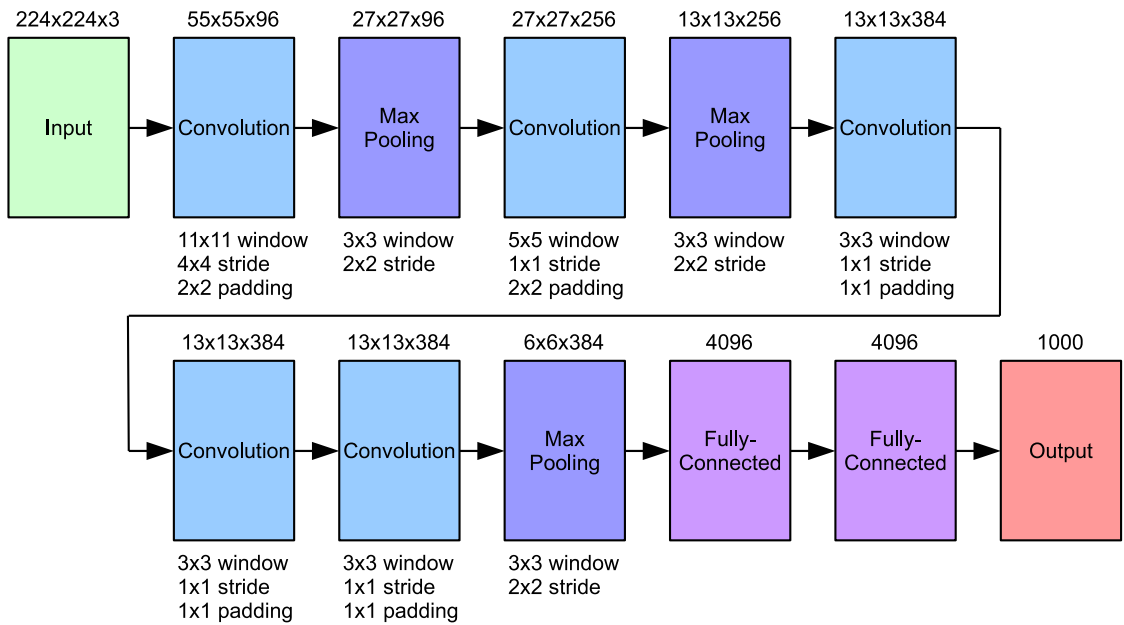}
\caption{The AlexNet, containing five convolutional , three pooling and two fully-connected layers between input and output layers}
\label{fig:alexnet}
\end{figure*}

The \emph{AlexNet}, developed by \cite{AlexNet}, is potentially the most influential implementation of DCNNs up to date. It was the first DCNN that managed to beat more traditional object recognition approaches in the ILSVRC. Moreover, the AlexNet proved the viability of DCNN approaches for object recognition tasks. The corresponding network architecture is shown in Fig. \ref{fig:alexnet}. As we can see, the AlexNet is not much different from the LeNet-, as it also consists of only the input layer, a few convolutional layers with occasional pooling afterward, as well as some fully-connected layers right before the output layer. However, the AlexNet has more layers and neurons per layer and it also uses different hyperparameters, as can be seen in Fig. \ref{fig:alexnet}.

\subsection{ZFNet and OverFeat (2013)}\label{subsec:zfnet}
One year after the AlexNet won the ILSVRC in 2012, a modification of it, the \emph{ZFNet}, still achieved best results, winning the ILSVRC again. \cite{ZFNet} developed a novel technique for visualizing convolutional neural networks, called \emph{Deconvolutional Network}. This network does the exact opposite of a CNN, mapping features to pixels, and is nowadays also frequently used in combination with CNNs for generative tasks \cite{GAN}. Visualizing the AlexNet with it enabled them to improve it by tuning its hyperparameters and increasing the number of filters in the later convolutional layers.

The winner of the 2013 ILSVRC localization challenge was an architecture named \emph{OverFeat} \cite{Overfeat}. OverFeat used the exact version of the AlexNet that won the classification challenge in 2012 and altered it in a novel way to perform the bounding box regression. Instead of only predicting bounding boxes once per image, as was suggested in Sect. \ref{subsec:localization}, OverFeat tried to localize a given object at multiple locations and scales and merges these outputs to obtain the final results.

\subsection{VGGNet and GoogLeNet (2014)}\label{subsec:2014}
\begin{figure*}
\includegraphics{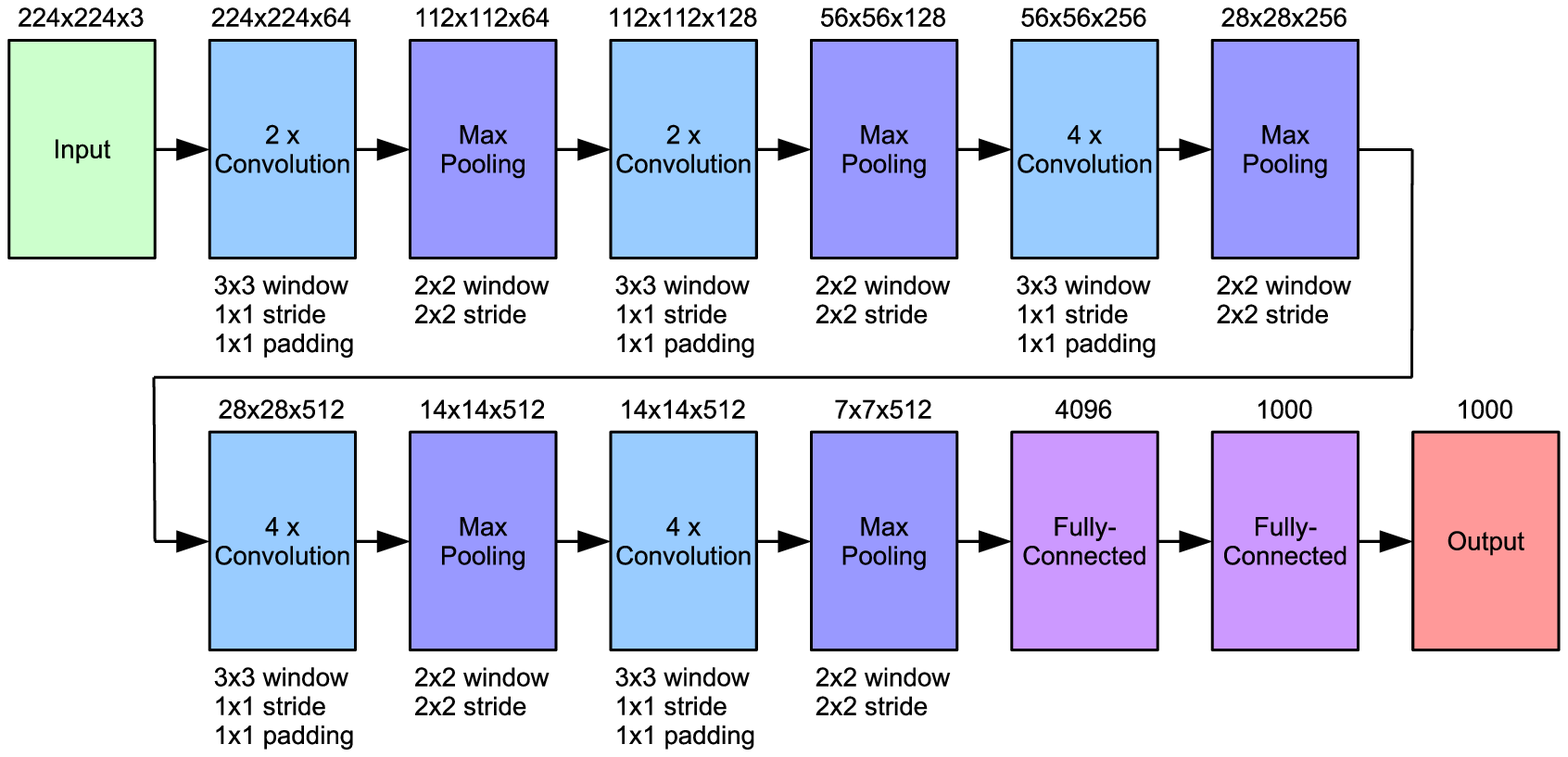}
\caption{The VGGNet, containing a total of 24 hidden layers, consisting of 16 $3\times3$ convolutional layers, five $2\times2$ max pooling layers, as well as three fully-connected layers}
\label{fig:vgg}
\end{figure*}

Another influential network was developed by \cite{VGG}. Their implementation, the \emph{VGGNet}, scored second place in the ILSVRC and influenced the deep learning scene in an important way, as they showed that using a deeper architecture does generally lead to better results, which was not obvious at that time. The VGGNet that was submitted for the ILSVRC contained $19$ parametrized hidden layers, which was much more than what previous architectures had used. Apart from its size, the VGGNet was very simple. It only consisted of convolutional layers with a  $3\times3$ receptive field, which is the smallest size that can differentiate basic directions, as well as $2\times2$ max pooling layers, and three fully-connected layers at the end. A scheme of the VGGNet is shown in Fig. \ref{fig:vgg}.

\begin{figure}
\includegraphics{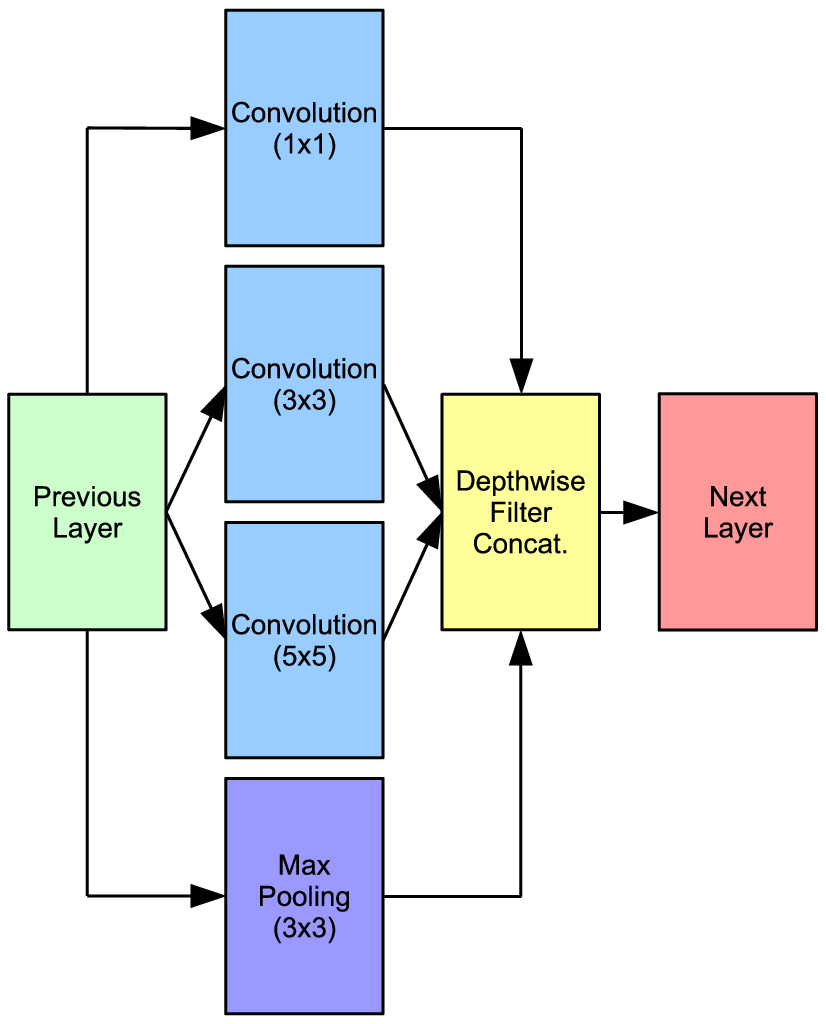}
\caption{An inception module, performing three convolutions with different receptive fields, as well as a pooling operation in parallel, as first presented by \cite{GoogLeNet}. The results of the different operations are combined by a depthwise filter concatenation to obtain the final output of the module}
\label{fig:inception_module}
\end{figure}

In the same year, Szegedy et al. from Google won the ILSVRC with a different very deep network that had $22$ parametrized layers - even more than the VGGNet - and was named \emph{GoogLeNet} \cite{GoogLeNet}. This network was an improvement of the AlexNet that was not only much deeper but also reduced the number of parameters. The latter was achieved by replacing the first fully-connected layer, which is typically accountable for the highest number of parameters, by another convolutional layer. In addition to that, they also implemented the so-called \emph{Inception Modules}, which enable a network to recognize patterns of different sizes within the same layer. In order to do so, the inception module performs several convolutions with different receptive fields in parallel and combines the results by merging the depth slices of the different filters into one single layer. Such an inception module can be seen in Fig. \ref{fig:inception_module}.

\begin{figure*}
\includegraphics{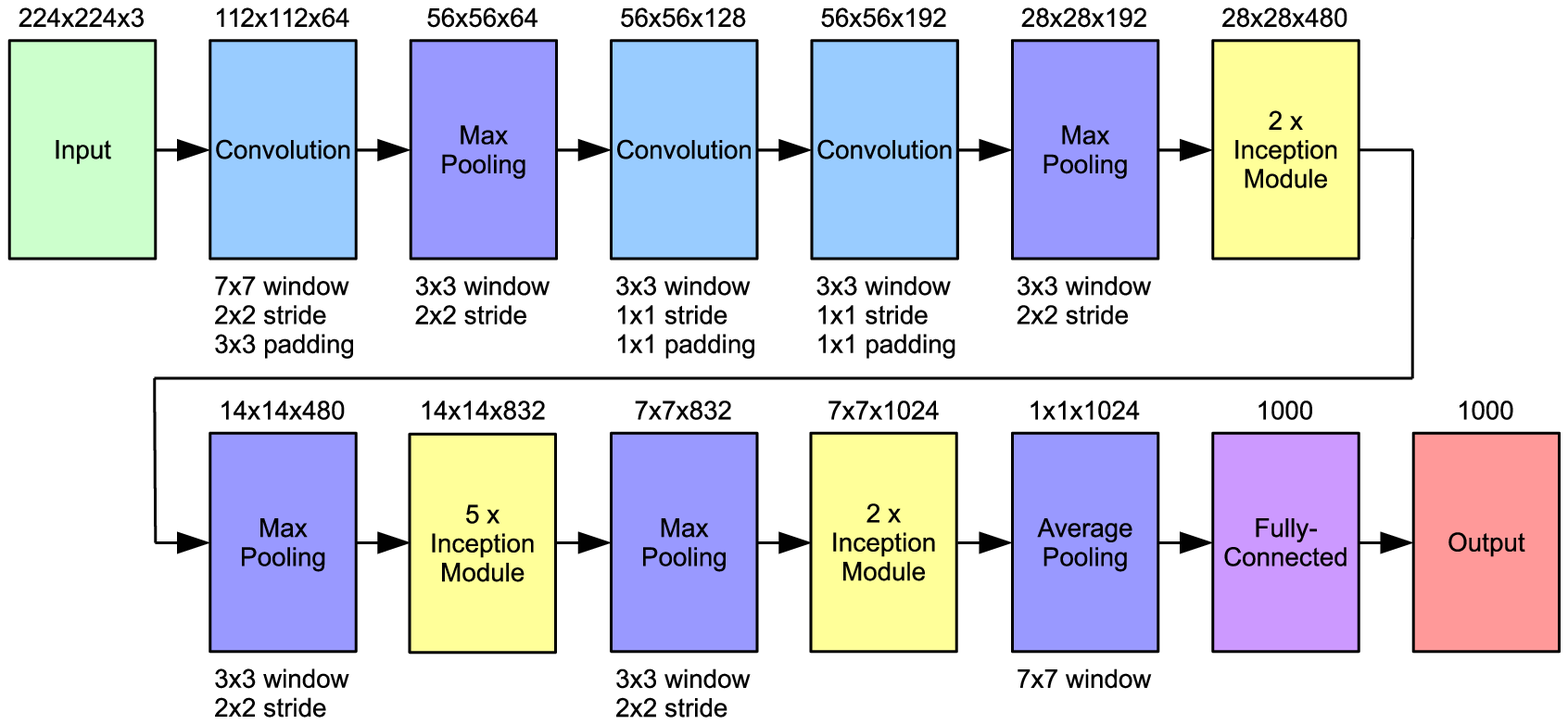}
\caption{The GoogLeNet, containing nine inception modules, five pooling layers, three convolutional layers after the input layer, as well as a fully-connected layer before the output layer}
\label{fig:googlenet}
\end{figure*}

The final GoogLeNet consisted of several such inception modules stacked on top of each other with occasional pooling layers in between, a few additional convolutional layers in the beginning of the network and a few fully-connected layers right before the output layer. This is shown in Fig. \ref{fig:googlenet}.

The GoogLeNet also contained additional output layers closer to the middle of the network and their outputs were combined with the output of the final layer of the network to obtain the total prediction. This had some minor influence on the overall result but was mainly intended to accelerate the training of earlier layers.

\subsection{ResNet (2015)}\label{subsec:2015}
\begin{figure}
\includegraphics{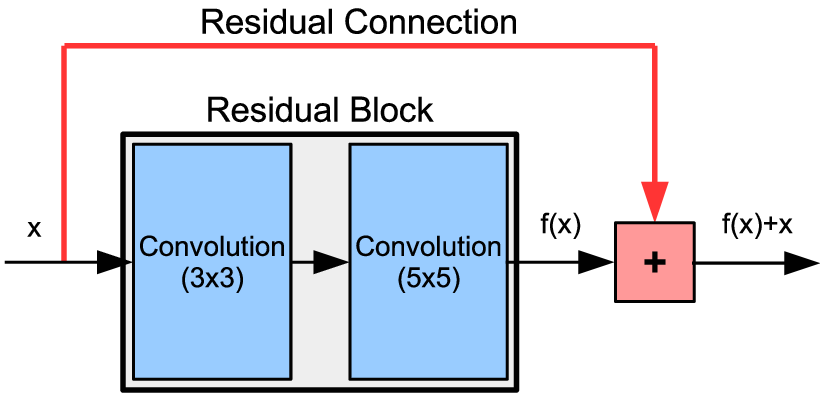}
\caption{A residual block, consisting of two convolutional layers, and the corresponding residual connection. The input to the residual block ($x$) is forwarded by the residual connection and later added to the output of the residual block ($f(x)$) to obtain the final output ($f(x)+x$)}
\label{fig:resnetblock}
\end{figure}

\cite{ResNet} developed a new kind of network architecture, which pushed the depth boundaries of DCNNs even further. Their network, called \emph{Deep Residual Network}, or \emph{ResNet} for short, is able to perform much better with very deep architectures. In ResNets, convolutional layers are divided into \emph{Residual Blocks} and for each block a \emph{Residual Connections} is added, which is bypassing the corresponding block. Afterwards, the output of the residual block is merged by summation with the original input that was forwarded by the residual connection, as can be seen in Fig. \ref{fig:resnetblock}. 

By adding these residual connections, the result of a training step can be backpropagated to the earlier layers directly, without any interference from subsequent layers. Therefore, residual connections enable the training of even deeper networks. Previously, having more layers did not guarantee a better accuracy, which was caused by earlier layers not adapting properly if a network was getting too big. As a result, He et al. won both the ILSVRC localization and classification contests, as well as the COCO detection and segmentation challenges \cite{COCO}. They also managed to improve on the previous error rates by a big margin. 

\begin{figure*}
\includegraphics{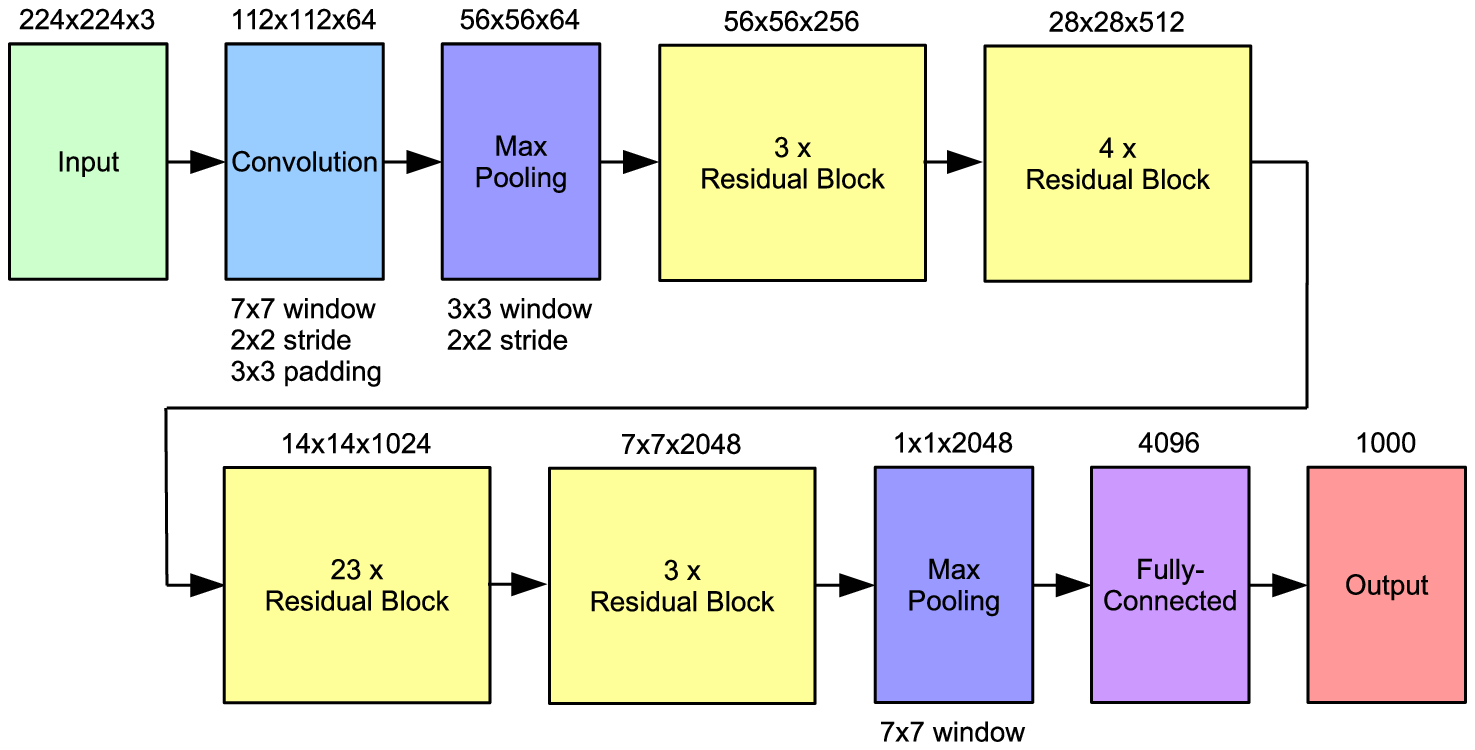}
\caption{The ResNet101, consisting of a $7 \times 7$ convolutional layer, a $2 \times 2$ max pooling layer, 33 residual blocks, a $7 \times 7$ average pooling layer and a final fully-connected layer before the output layer}
\label{fig:resnet101}
\end{figure*}

The ResNet version that was submitted to these contests was the \emph{Resnet101}, which consists of $101$ parametrized layers. These $101$ layers consist of an initial $7 \times 7$ convolutional layer with a $2 \times 2$ stride, $33$ residual building blocks with decreasing output size and increasing depth and a final $1000$ neuron fully-connected layer  \cite{ResNet}. In addition to the $101$ parametrized layers, the ResNet101 also includes one max pooling layer after the first convolutional layer and an average pooling layer before the final fully-connected layer. The overall structure of the ResNet101 is shown in Fig. \ref{fig:resnet101}. Note that the first convolutional layer of each group of blocks uses stride two in order to achieve the output size reduction.

\subsection{Inception-v4, Inception-ResNet-v1 and ResNeXt (2016)}\label{subsec:2016}
Early in 2016, a new modification of the GoogLeNet has been released by \cite{Inception}. This network, called \emph{Inception-v4}, is the fourth iteration of the GoogLeNet and consisted of many more layers than the original version. During the continuous improvements on the inception architectures, the inception modules, as introduced in Sect. \ref{subsec:2014}, have been vastly improved as well and the Inception-v4 uses three different kinds of inception modules. 

\begin{figure}
\includegraphics{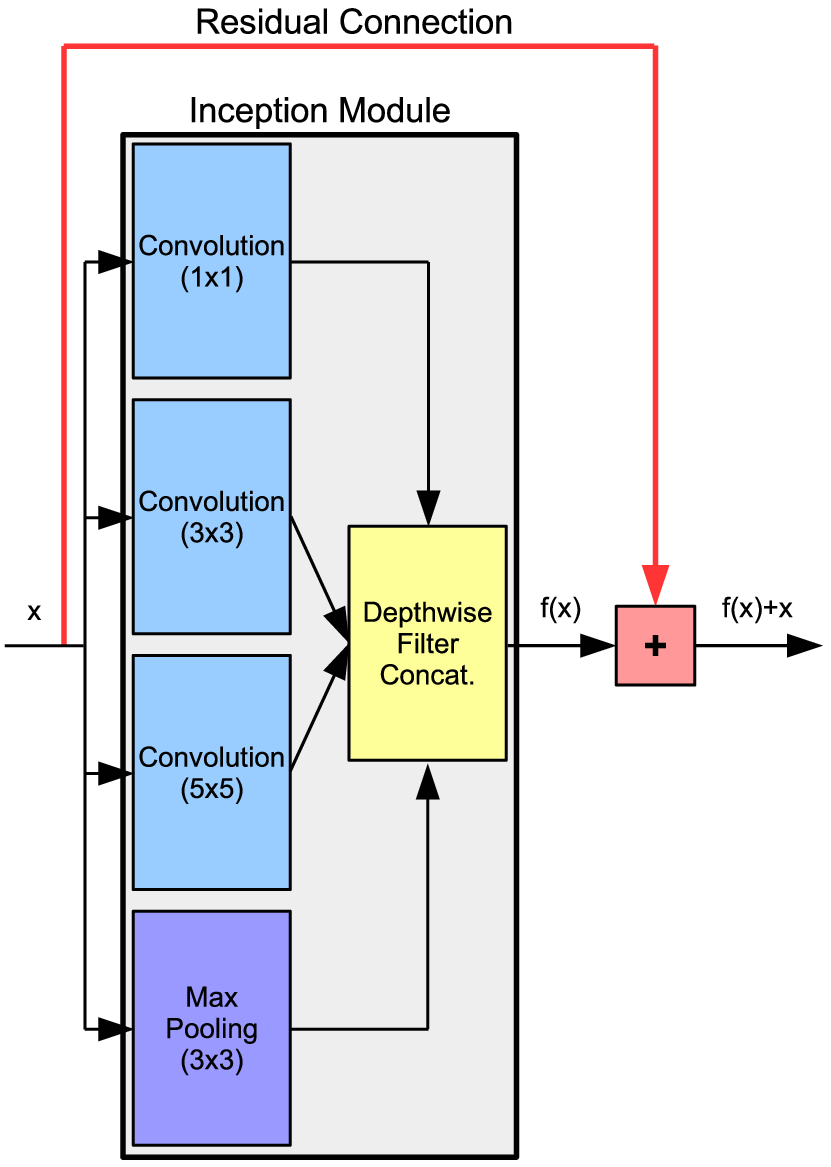}
\caption{A building block of an Inception-ResNet. The input (x) is processed by an inception module and also forwarded by a residual connection}
\label{fig:inceptionresnet}
\end{figure}

In addition to the Inception-v4, the corresponding paper also introduced a new type of network, named \emph{Inception-ResNet}, which is a combination of an inception network and a ResNet, by combining the inception module and residual connection as shown in Fig. \ref{fig:inceptionresnet}.

This makes the network even more efficient, leading to much lower training times compared to a similarly complex inception network. Both of these network architectures are hundreds of layers deep and contain a wide variety of layers, inception modules and residual blocks, for which reason we will not inspect their structures in more detail \footnote{The exact network architecture is explained in detail in the original paper by \cite{Inception}.}.

\begin{figure}
\includegraphics{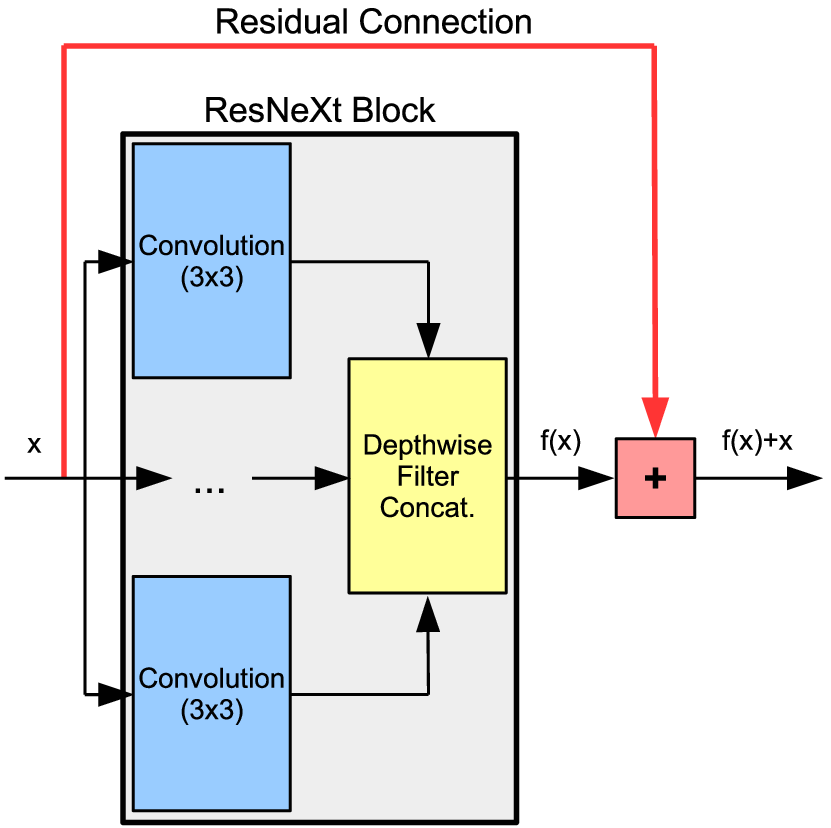}
\caption{A residual block of a ResNeXt. Within the block, multiple convolutions with similar hyperparameters are performed in parallel and merged by filter concatenation}
\label{fig:resnext}
\end{figure}

The second place in the 2016 ILSVRC classification challenge was achieved by \cite{ResNext}, who proposed a new network architecture, named \emph{ResNeXt}. Similarly to Inception-ResNets, the ResNeXt architecture is combining the residual building blocks and connections of ResNets with the parallelization strategy of inception architectures. Unlike Inception-ResNets, the different convolution paths that are concatenated in ResNeXts have similar hyperparameters, as shown in Fig. \ref{fig:resnext}.

Therefore, the number of paths is variable and the paths do not have to be adapted for a specific purpose, which facilitates the design of new ResNext variants. Furthermore, the flexibility of the design enables ResNext to introduce a new hyperparameter called \emph{Cardinality}, which specifies how many parallel convolution paths each block contains. According to \cite{ResNext}, increasing the cardinality of a network has a stronger influence on the performance of a network than increasing the number of layers or the number of neurons per layer, for which reason their ResNext managed to outperform all previous Inception-ResNet architectures while having a much simpler design, as well as a lower complexity.

The best performing approaches on the 2016 ILSVRC classification, localization and detection challenges all used ensembles \footnote{Network ensembles will be inspected more closely in the next section.} of ResNet101, Inception-v4 and Inception-ResNet-v1 networks.

\subsection{Densenet, DPN and MobileNets (2017)}\label{subsec:2017}

\begin{figure*}
\includegraphics{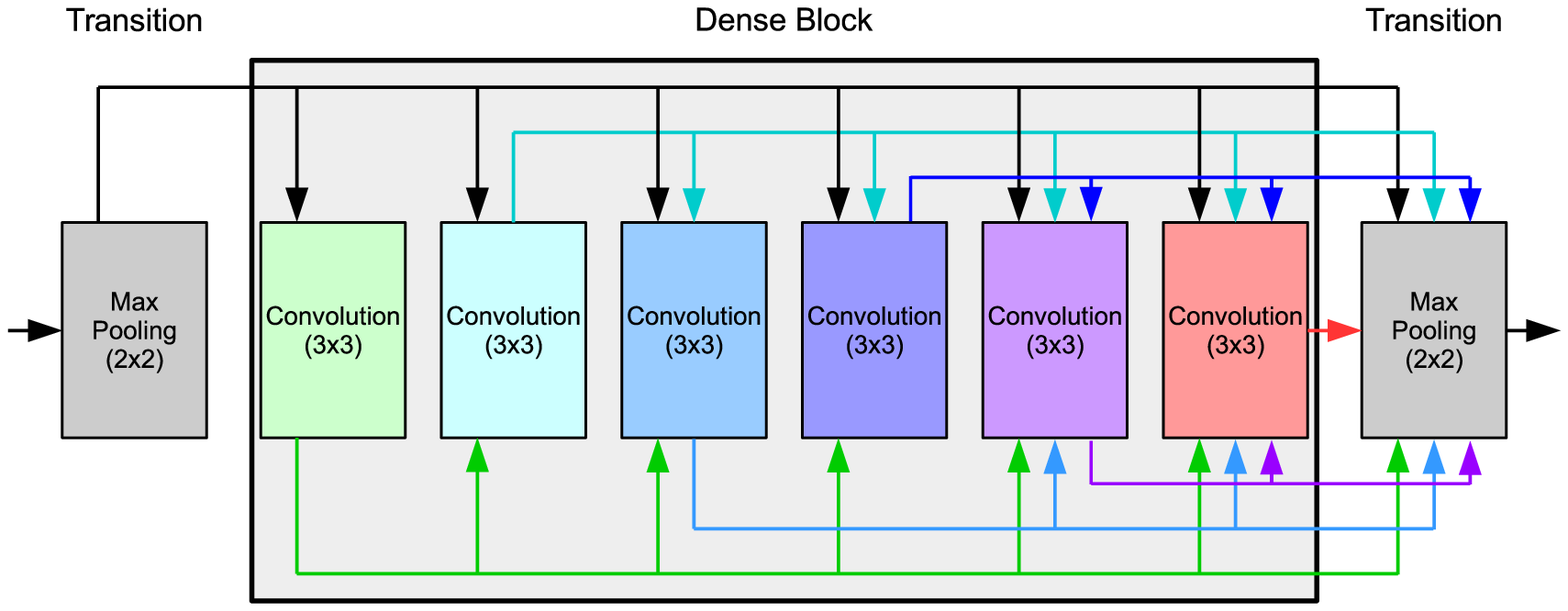}
\caption{A dense block of a DenseNet together with the previous and subsequent transition layers. Every layer is directly connected to every subsequent one}
\label{fig:densenet}
\end{figure*}

As described in Sect. \ref{subsec:2015}, the enhanced information flow of residual connections in ResNets and its modifications is enabling the training of much deeper networks. Instead of summing up the output of a residual connection with the output of the corresponding residual block, as shown in Fig. \ref{fig:resnetblock}, \emph{Dense Convolutional Networks (DenseNets)}  \cite{DenseNet} combine the two outputs by depthwise filter concatenation, as performed in inception modules. Furthermore, DenseNets are adding one such connection from each layer to all subsequent ones with matching input sizes. By doing so, the learned features of a layer can be reused by any of the following layers. Therefore, later layers need to produce much fewer feature maps, resulting in less complex architectures with fewer parameters. Since the width and height of layers in CNNs are gradually decreasing, connecting all compatible layers is dividing the network into \emph{Dense Blocks}. Between these blocks, pooling layers are used to alter the sizes accordingly. These layers are referred to as \emph{Transition Layers}. An example of a dense block with previous and subsequent transition layers is shown in Fig. \ref{fig:densenet}.

Due to the high layer interconnectivity, DenseNets are easy to train and naturally scale well with increasing depth and increasing amount of parameters.

\begin{figure*}
\includegraphics{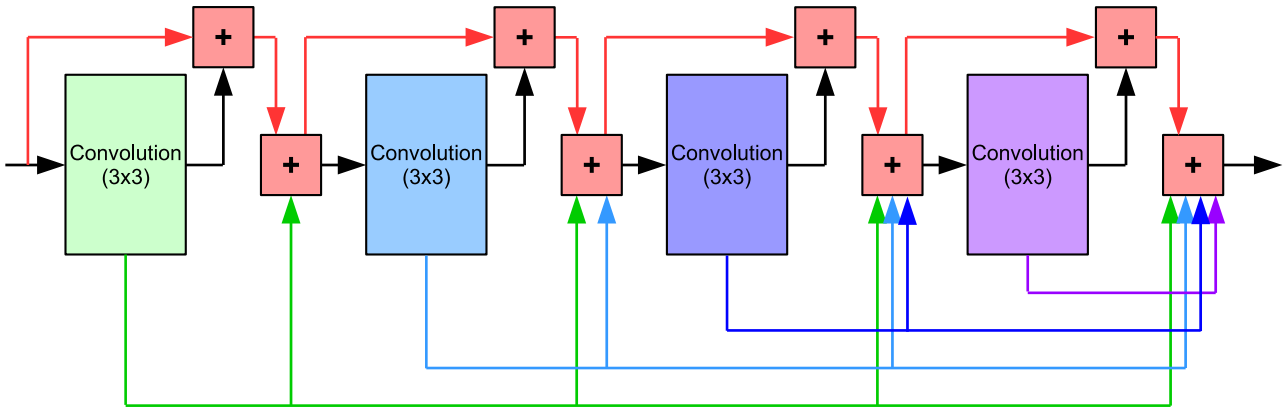}
\caption{Four convolutional layers in a block of a DPN. Each layer has a corresponding residual connection and is also forwarding its output to all subsequent layers. Before each layer, the two streams of information are merged by performing addition}
\label{fig:dpn}
\end{figure*}

According to \cite{DPN}, the design of DenseNets encourages the learning of new features, while ResNet architectures is leading to increased feature reuse. Since both architectures have advantages over each other, Chen et al. combined the two architectures into a \emph{Dual Path Network (DPN)}, with which they won first place in the 2017 ILSVRC localization challenge and finished top three in both classification and detection. In order to combine the networks, the output of a layer is split and one part is combined with a residual connection, whereas the other is forwarded to all subsequent layers, as performed in DenseNets. This approach is illustrated in Fig. \ref{fig:dpn}.

\begin{figure*}
\includegraphics{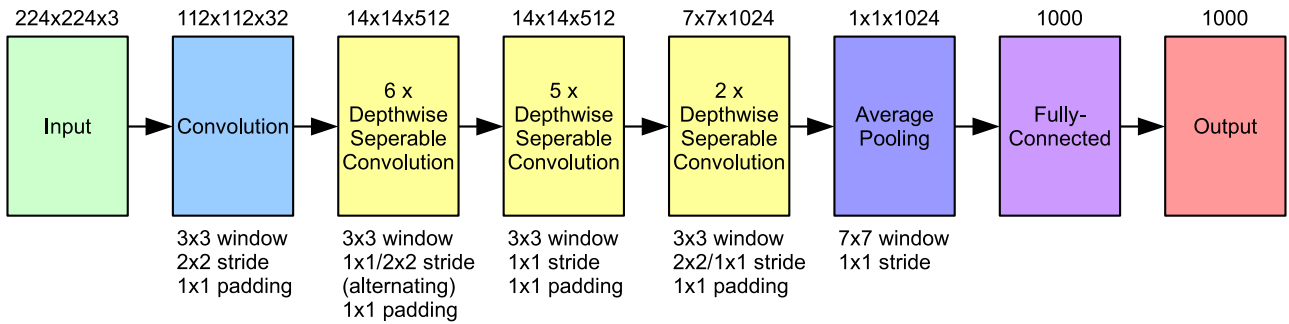}
\caption{The MobileNet-v1, consisting of a $3 \times 3$ convolutional layer, 13 $3 \times 3$ depthwise separable convolutions, a $7 \times 7$ average pooling layer and a final fully-connected layer before the output layer}
\label{fig:mobilenetv1}
\end{figure*}

All of the previous architectures we inspected were designed to achieve the highest accuracies possible, for which reason recent architectures have become ever more complex. While a high complexity is not problematic for challenges like the ILSVRC, it is very cumbersome for real-time applications with restricted hardware, such as mobile applications or embedded systems. Therefore, a new DCNN architecture was introduced to tackle this problem. These networks, called \emph{MobileNets} \cite{MobileNets}, are built to be as time efficient as possible by replacing standard convolutions by \emph{Depthwise Separable Convolutions}, as first introduced by \cite{DSConv}. As we have seen in Sect. \ref{subsubsec:convolution}, a neuron in a convolutional layer combines the outputs of a square sized region across the width and height dimensions and across the whole input depth. Depthwise separable convolutions split this process into two steps: A \emph{Depthwise Convolution} and a \emph{Pointwise Convolution}. The depthwise convolution acts as a filter by only considering the square sized regions within a single depth slice. The point wise convolution then performs a $1\times1$ convolution to merge the information across the whole depth. Additionally, two new hyperparameters are used for MobileNets, which enable creations of even faster architectures by trading off accuracy and execution time. The first hyperparameter, called \emph{Width Multiplier}, is a value between zero and one that lowers the number of neurons in all layers by the given factor, which results in a squared reduction of execution time and has been found to decrease the output quality less severely than lowering the number of layers in the network \cite{MobileNets}. The other hyperparameter, \emph{Resolution Multiplier}, lowers the resolution of the input image by the given factor, which also results in a squared reduction of execution time. By adjusting these parameters it is, therefore, possible to construct MobileNets that exactly match the execution time requirement of a given application while preserving a relatively high quality of results. The first version of such a MobileNet is shown in Fig. \ref{fig:mobilenetv1}.

\begin{figure*}[h!]
\includegraphics{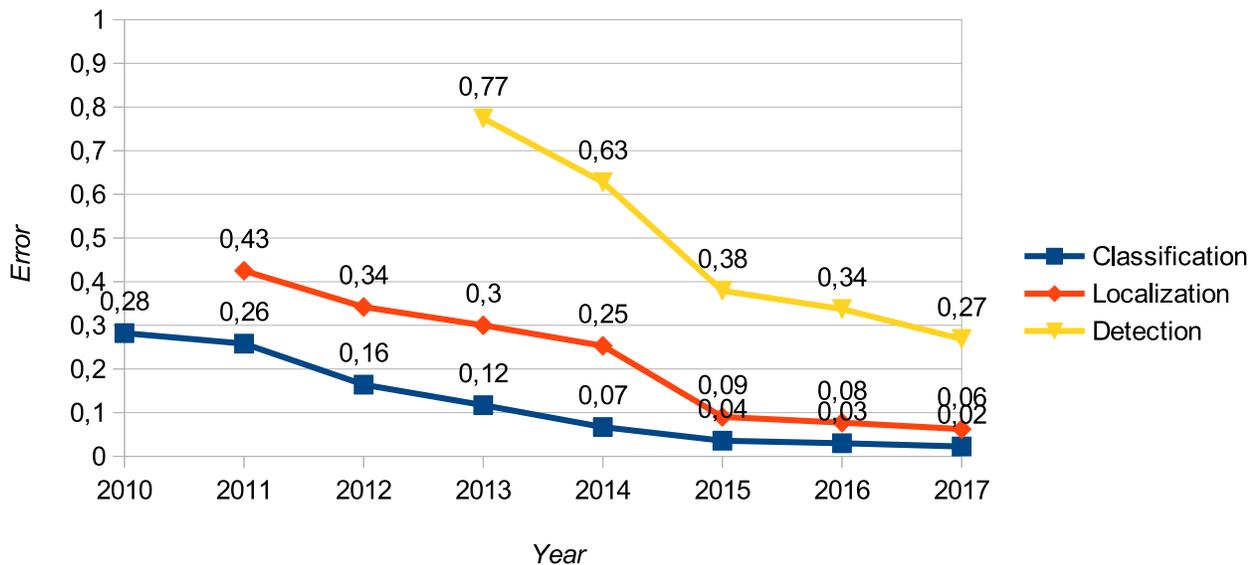}
\caption{Results of the best performing architectures on the ILSVRC classification, localization and detection challenges}
\label{fig:improvement}
\end{figure*}

\subsection{Improvement over time}
After having inspected the best architectures of each year, it would be interesting to see, how effective each innovation and modification was and how fast the field is advancing. For this reason, we visualized the results of the all winning architectures on the classification, localization and detection tasks in the ILSVRC in Fig. \ref{fig:improvement}. Since detection accuracy is measured in \emph{Mean Average Precision (MAP)}, we define the corresponding error as $error = 1-MAP$.

As Fig. \ref{fig:improvement} is showing, the classification error has been decreased by a large quantity due to DCNNs. In 2011, when the winning architecture was not a DCNN yet, the classification error amounted to $26$ percent and only five years later, in 2016, it was possible to lower the error to three percent, which is even lower than the human error rate of about five percent \cite{ILSVRC}. The localization and detection results have been steadily improving as well, but as we can see, the advent of residual networks in 2015 had a particularly strong impact.

\section{Conclusion}\label{sec:conclusion}
In this paper, we have explored what deep convolutional neural networks are, why they work as well as they do for object recognition tasks and how current state-of-the-art architectures are composed. As suggested in \ref{subsec:classification}, most of the DCNN architectures we inspected follow a clear two-part design pattern: In the first part (at the beginning of the network) multiple convolution and pooling layers are stacked on top of each other to produce abstract data representations and in the second part (at the end of the network) additional fully connected layers are used to forward these abstractions to the output layer. Regarding hyperparameters, a good choice for the convolution layer seems to be a window size of three with a padding of one and a stride of one or two. For pooling layers, a window size of two with no padding and a stride of one or two are frequently selected. Finally, in fully-connected layers, the amount of neurons should be greater or equal to the amount of neurons in the following layer and be within the same order of magnitude. One of the more general key design ideas that we can observe in recent network architectures is that direct connections from earlier to later layers seem to be necessary to achieve state-of-the-art results. As explained in Section \ref{subsec:2015}, such connections are crucial for networks with many layers, as the information on the prediction quality can otherwise not be reliably transmitted back to earlier layers. Another key design idea is that intra-layer parallelism, as seen in Inception architectures and ResNext, is desirable in large networks. As mentioned in \ref{subsec:2016}, such parallel convolution paths significantly decrease the network complexity, as it leads to less layers and less neurons per layer being required to achieve similar results.

Lastly, we would like to mention some selected resources for further research. To readers that are interested in a more in-depth introduction to deep convolutional neural networks, we highly recommend the Stanford University course CS231n\footnote{The lecture material and assignments of the CS231n course can be found online at http://cs231n.github.io/ and recordings of the lecture are available on YouTube.}. For further details about specific parts of this survey, we strongly recommend to consult the corresponding original papers as listed in the \emph{References} section.

\section*{Acknowledgment}
The first version of this paper was created during the seminar \emph{Human-Robot-Interaction}, held in a joint cooperation by the \emph{Technical University of Munich} and \emph{Fortiss GmbH}. Therefore, the authors would like to thank both corporations and the supervisor of the paper, Andrej Pangercic, in particular. Furthermore, we would like to thank Oleksandr Melkonyan, Yuriy Arabskyy and Cristian Plop for their valuable inputs and corrections.

\bibliographystyle{IEEEtran}
\bibliography{IEEEabrv,bibliography}
\end{document}